\documentclass[runningheads]{llncs}
\usepackage{booktabs}
\usepackage{multirow}
\usepackage[T1]{fontenc}
\usepackage{amsmath}
\usepackage{hyperref}

\usepackage{graphicx,verbatim}
\usepackage{color}

\begin{document}
\title{MR-CLIP: Efficient Metadata-Guided Learning of MRI Contrast Representations}

\author{Mehmet Yigit Avci\inst{1}, Pedro Borges\inst{1}, Paul Wright\inst{1}, Mehmet Yigitsoy\inst{2}, Sebastien Ourselin\inst{1}, and Jorge Cardoso\inst{1}}
\authorrunning{M.Y. Avci et al.}
%
\institute{School of Biomedical Engineering and Imaging Sciences, King’s College
London, London, UK \and deepc GMBH, Munich, Germany}

\titlerunning{MR-CLIP}
  
\maketitle              
\begin{abstract}
Accurate interpretation of Magnetic Resonance Imaging scans in clinical systems is based on a precise understanding of image contrast. This contrast is primarily governed by acquisition parameters, such as echo time and repetition time, which are stored in the DICOM metadata. To simplify contrast identification, broad labels such as T1-weighted or T2-weighted are commonly used, but these offer only a coarse approximation of the underlying acquisition settings. In many real-world datasets, such labels are entirely missing, leaving raw acquisition parameters as the only indicators of contrast. Adding to this challenge, the available metadata is often incomplete, noisy, or inconsistent. The lack of reliable and standardized metadata complicates tasks such as image interpretation, retrieval, and integration into clinical workflows. Furthermore, robust contrast-aware representations are essential to enable more advanced clinical applications, such as achieving modality-invariant representations and data harmonization. To address these challenges, we propose MR-CLIP, a multimodal contrastive learning framework that aligns MR images with their DICOM metadata to learn contrast-aware representations, without relying on manual labels. Trained on a diverse clinical dataset that spans various scanners and protocols, MR-CLIP captures contrast variations across acquisitions and within scans, enabling anatomy-invariant representations. We demonstrate its effectiveness in cross-modal retrieval and contrast classification, highlighting its scalability and potential for further clinical applications. The code and weights are publicly available at \url{https://github.com/myigitavci/MR-CLIP}.


\end{abstract}
\section{Introduction}
Magnetic resonance imaging (MRI) is the cornerstone of modern neuroimaging, offering diverse contrast mechanisms sensitive to different tissue properties and pathologies. However, in clinical settings, the interpretation and organization of MR images is challenged by substantial variability in acquisition protocols, scanner vendors, and data completeness \cite{sinha2024mrqa}. Unlike curated research datasets that are relatively homogeneous, real-world clinical data is highly variable and often lack explicit sequence labels such as T1-weighted or T2-weighted \cite{gauriau2020dicom}. Even when present, these labels offer only a coarse approximation of image contrast. Instead, the true visual characteristics of an MR image are determined by acquisition settings, such as scanner type, echo time, and repetition time, which are embedded within the Digital Imaging and Communications in Medicine (DICOM) \cite{DICOMStandard} metadata. Reliable and complete metadata are essential for clinical interpretation, image retrieval, and picture archiving and communication systems (PACS) routing and archiving. Furthermore, robust contrast-aware representations are foundational for clinical applications such as cross-site and cross-modality data harmonization, and modality-invariant representations \cite{haca3,ouyang2021representation}.

Recent studies highlight the importance of metadata in shaping MR image appearance; for example,  \cite{contextmri} incorporates DICOM metadata into image reconstruction, showing that contrast-aware models can improve image quality. Similarly, a contrastive learning approach for retinal OCT images  uses patient metadata to enhance representations \cite{oct-clip}. TUMSyn \cite{tumsyn} synthesizes MR images guided by DICOM tags using contrastive learning; however, it relies primarily on controlled research datasets with limited scanner and protocol diversity. Other works~\cite{mcdaniel2022contrast,du2020radnet,gauriau2020dicom,liang2021mri} use supervised learning to estimate metadata or sequence type from images or tags, though they mainly predict high-level labels such as T1/T2 contrast while overlooking fine-grained MR sequences parameter variability. CLIP~\cite{radford2021learning} learns aligned image–text representations via contrastive learning, enabling powerful zero-shot transfer across diverse tasks. In medicine, models such as BiomedCLIP~\cite{biomedclip}, PMC-CLIP~\cite{wu2023pmcclip}, and MedCLIP~\cite{wang2022medclip} adapt this framework to medical data for retrieval and classification. However, they primarily focus on general tasks, overlooking domain-specific MR acquisition metadata that critically shapes image contrast and appearance.

In this work, we present MR-CLIP, a scalable multimodal contrastive learning framework that aligns MR images with their corresponding DICOM metadata to learn generalizable MR contrast representations. Unlike prior methods that depend on expert annotations or curated datasets, MR-CLIP is trained on a large, diverse hospital dataset with all types of brain MRIs, spanning wide range of acquisition protocols and scanner variations. Our contributions are as follows:
\begin{itemize}
\item We introduce a scalable multimodal contrastive learning approach that effectively utilizes DICOM metadata to supervise the learning process, eliminating manual contrast labeling or protocol standardization.
\item We extend Supervised Contrastive Loss (SupCon) \cite{supcon} to operate both across scans and within 3D volumes, capturing fine-grained contrast variations while promoting invariance to anatomy.
\item We design a metadata grouping strategy to enhance the stability and generalizability of supervision for contrastive learning.
\item We validate MR-CLIP on a diverse clinical dataset and demonstrate its effectiveness on cross-modal retrieval and contrast classification. We analyze failure cases by identifying frequently confused DICOM tags and demonstrate MR-CLIP's generalization to an unseen dataset, highlighting its robustness.
\end{itemize}

\begin{figure*}[tb!]
\centering
\includegraphics[width=0.99\textwidth]{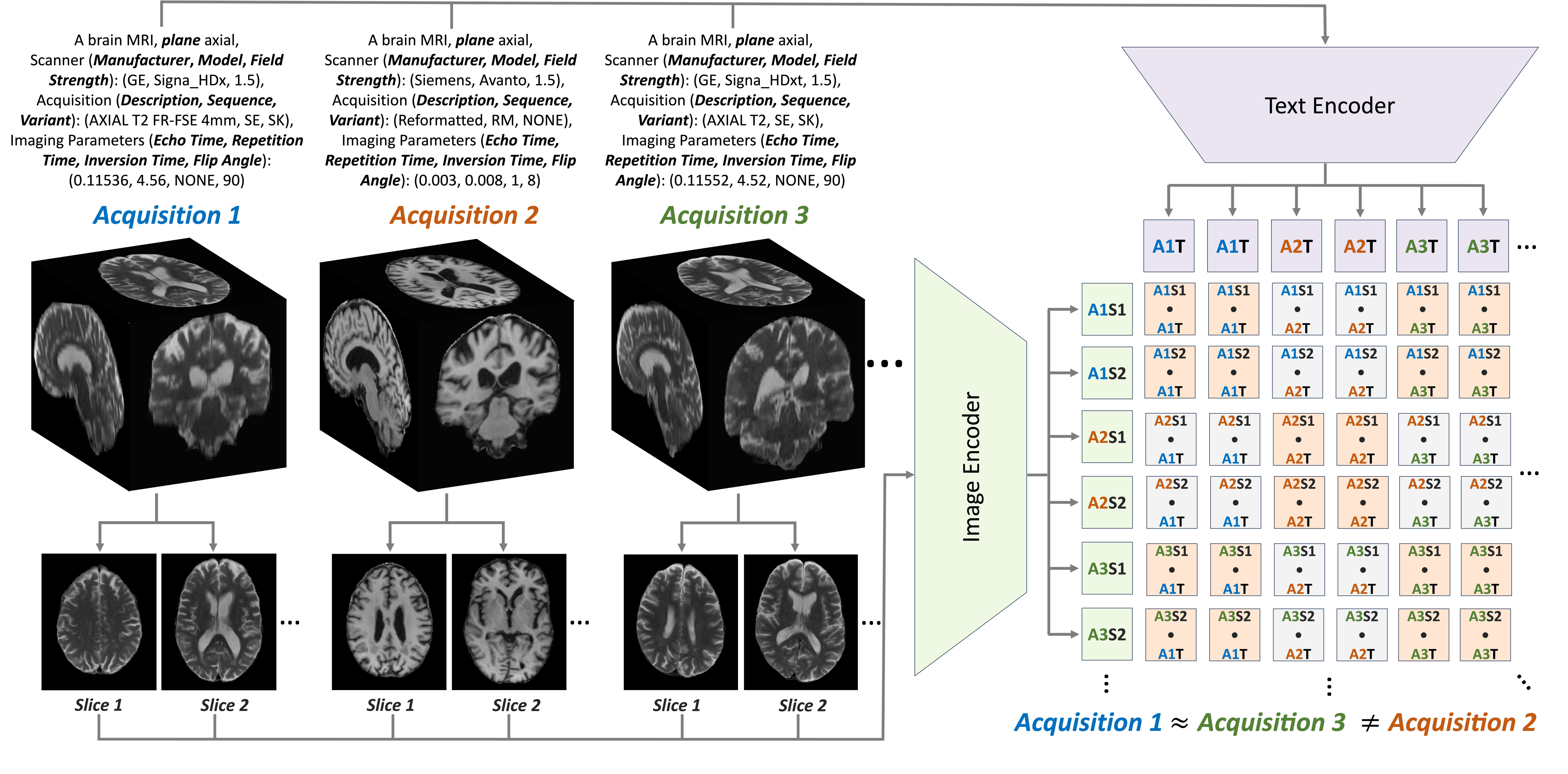}
\caption{MR-CLIP leverages slice-level image and metadata pairs across multiple acquisitions and slices. An image encoder processes slices, while a text encoder embeds corresponding structured DICOM metadata, with MR-CLIP contrastively learning to align image and metadata representations. Representations of slices of similar acquisitions (Acq. 1 and Acq. 3) are pulled together, while those of distinct contrasts (Acq. 2) are pushed apart, enabling the model to learn fine-grained, contrast-aware features that are invariant to anatomy and subtle acquisition variability.}
\label{fig1}
\end{figure*}
\section{Methods}

The MR-CLIP framework is designed to learn robust, contrast-aware representations by aligning 2D MR slices and their corresponding metadata into a shared embedding space using contrastive learning as illustrated in Fig.~\ref{fig1}. A vision encoder embeds the images, while a text encoder embeds structured metadata. By grouping embeddings that share similar contrast characteristics, MR-CLIP enables generalization across acquisition settings and anatomical variations. 

\begin{figure*}[tb!]
\centering
\includegraphics[width=0.91\textwidth]{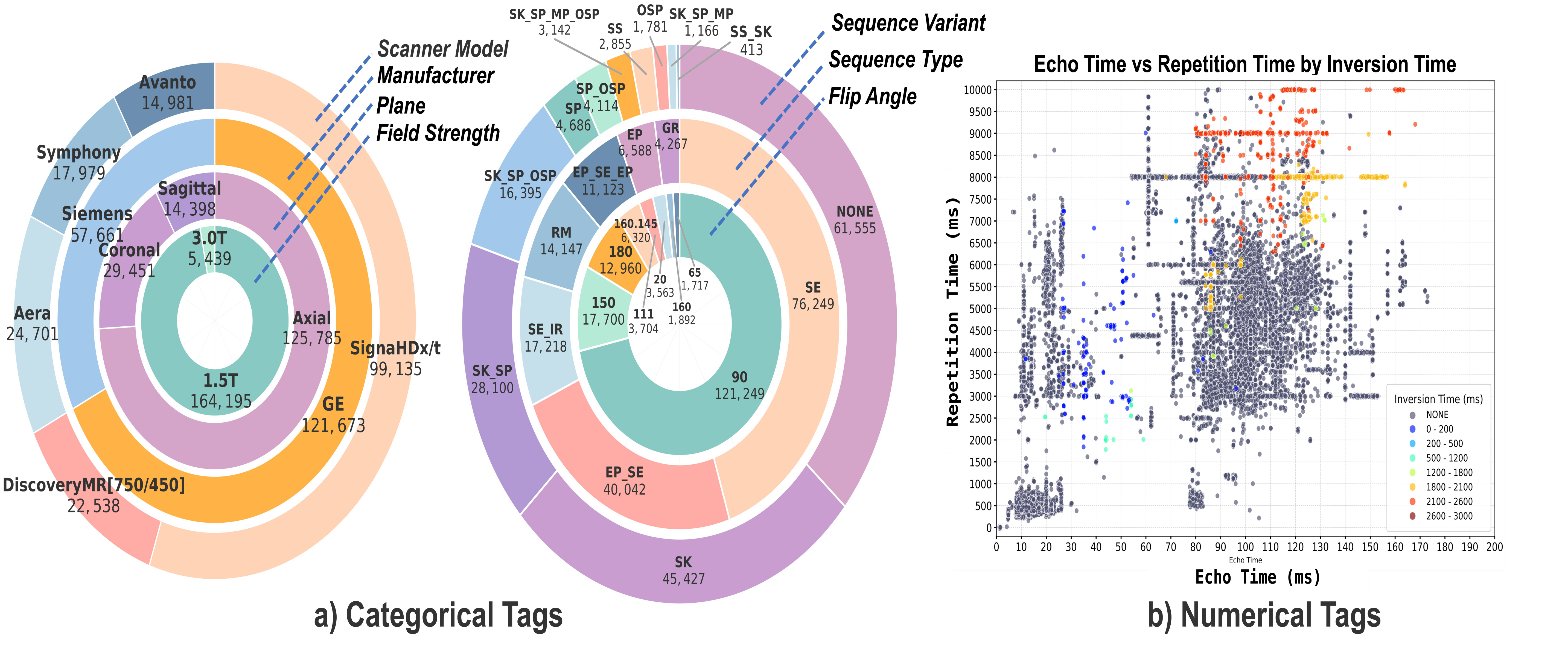}
\caption{Overview of metadata distribution in our dataset. (a) Categorical tags including scanner, plane, field strength, and sequence information and flip angle. (b) Numerical distribution of echo and repetition times, color-coded by inversion time.}
\label{fig2}
\end{figure*}

To guide alignment between MR images and metadata without manual labeling, a structured label space is constructed using critical DICOM metadata fields that jointly influence image contrast. These fields include categorical tags, including \textit{Manufacturer}, \textit{Scanner Model}, \textit{Imaging Plane}, \textit{Field Strength}, \textit{Sequence Type}, \textit{Sequence Variant}, \textit{Series Description}, \textit{Flip Angle}, and numerical tags \textit{Echo Time (TE)}, \textit{Repetition Time (TR)}, and \textit{Inversion Time (TI)}. Fig.~\ref{fig2} illustrates the distributions of the metadata fields. Each contrast-aware label corresponds to a unique combination of these metadata values. Categorical tags (Fig.~\ref{fig2}a) are grouped by exact values. Numerical tags are discretised to reduce the sensitivity to minor acquisition differences that do not meaningfully affect the image appearance (Fig.~\ref{fig2}b). TE and TR are jointly quantized into a $20\times20$ grid covering typical ranges observed in clinical protocols, with TI binned separately as >85\% scans lack inversion pulses and the remainder follow a sparse but structured distribution. While the \textit{Series Description} field can offer useful context, it is excluded from label construction due to site-dependent variability; instead, it is used only as an auxiliary input. 

Unlike traditional CLIP, which uses one-to-one image-text pairs with an InfoNCE loss \cite{infonce}, MR-CLIP employs supervised contrastive (SupCon) loss~\cite{supcon} that supports many-to-many positives. This is well-suited to MRI, where small variations in acquisition parameters (e.g., \textit{TE}, \textit{TR}) do not affect the underlying contrast. While CLIP would penalize these near-identical cases due to strict pairings, SupCon treats all samples sharing the same grouped contrast label as positives, enabling more flexible and semantically consistent supervision. Moreover, by aligning slices from different subjects and anatomical locations, MR-CLIP aims to learn contrast representations that are invariant to anatomy and subject-specific variation. Formally, given a batch of embeddings ${z_i}_{i=1}^N$, the SupCon loss for an anchor embedding $z_i$ is defined as:
\begin{equation}
\mathcal{L}_{\text{SupCon}} = \sum_{i \in \mathcal{I}} \frac{-1}{|P(i)|} \sum_{p \in P(i)} \log \frac{\exp(z_i \cdot z_p / \tau)}{\sum_{a \in \mathcal{A}(i)} \exp(z_i \cdot z_a / \tau)},
\end{equation}
where $P(i)$ is the set of all other embeddings in the batch that share the same grouped contrast label as anchor $i$, $\mathcal{A}(i)$ is the set of all other embeddings excluding $i$, and $\tau$ is a temperature parameter. Final loss is given as follows:
\begin{equation}
\mathcal{L} = \frac{1}{2} \left( \mathcal{L}_{\text{SupCon}}^{\text{img} \rightarrow \text{text}} + \mathcal{L}_{\text{SupCon}}^{\text{text} \rightarrow \text{img}} \right),
\end{equation}

\textbf{Dataset:} 
A large-scale dataset of 3D brain scans from \textit{King’s College Hospital (KCH)} and \textit{Guy’s and St Thomas’ NHS Foundation Trust (GSTT)} is used.
The dataset includes 40,005 subjects and 169,634 3D volumes, with 21,660 unique acquisition configurations derived from DICOM metadata. After applying our grouping strategy, these are consolidated into 1,415 unique contrast-aware labels. We split the dataset into training (60\%), validation (10\%), and test (30\%) sets at the scan level.

\textbf{Data Preprocessing:} All 3D volumes are first rigidly registered to the MNI template space and skull-stripped using the SynthStrip \cite{synthstrip}. From each registered volume, we extract a representative subset of slices by selecting every second slice from the middle 100 slices, which typically captures the most diagnostically relevant anatomical content while controlling dataset size. For plane determination (axial, coronal, sagittal), we infer acquisition orientation based on voxel resolution, selecting the highest resolution dimension as the slicing axis; if the image is originally isotropic, the axial plane is chosen. Inspired by \cite{tumsyn}, we convert metadata into natural language prompts using a standardized sentence template applied uniformly across the dataset to be readily interpretable by language encoders as can be seen in Fig. \ref{fig1}.

\textbf{Implementation Details:} MR-CLIP is implemented in PyTorch and trained on 3 NVIDIA A100 GPUs (40GB), using a batch size of 3000 for each GPU with sharding loss following CLIP implementation \cite{open_clip}. We optimize with Adam ($\text{lr}=1\mathrm{e}{-4}$, $\beta_1=0.9$, $\beta_2=0.98$) and a weight decay of 0.2, training for 100 epochs with 2000 warm-up steps. Gradient checkpointing is enabled to reduce memory usage. Patch dropout (0.5) and text dropout (0.2) are applied, along with standard image augmentations (random affine, resized crop, Gaussian blur, horizontal flip).

\section{Experiments}


\textbf{Cross-modal Retrieval and Linear Classification:} We investigate the impact of loss functions and architectural choices on representation quality, and additionally compare MR‑CLIP to BiomedCLIP, a vision–language model pre-trained on large-scale biomedical image–text pairs, in both its original frozen and fully fine-tuned configurations, as summarized in Table~\ref{tab:model_retrieval_comparison}. We evaluate models on two tasks: (i) cross-modal retrieval across three settings: retrieving metadata from 2D slices (image-to-text), from 3D scans via majority voting across slices (3D scan-to-text), and retrieving images from metadata (text-to-image); and (ii) linear probe classification, where a simple linear classifier is trained on frozen image embeddings to predict metadata labels. MR-CLIP achieves the highest (R@1) in both 3D scan-to-text (78.7\%) and text-to-image (90.9\%) retrieval, significantly outperforming the the fine-tuned BiomedCLIP (67.4\% R@1 for 3D scan-to-text). While InfoNCE achieves higher image-to-text R@5 and R@10 scores, likely due to its one-to-one training supervision, MR-CLIP demonstrates the strongest overall performance. Furthermore, MR-CLIP attains the highest linear classification accuracy (82.6\%), indicating that its learned representations are more discriminative and better aligned with metadata.

\begin{table*}[b]
\centering
\caption{Cross-modal retrieval performance (\%). Showing R@1/5/10 for image-to-text, 3D scan-to-text, and text-to-image retrieval. Linear classification accuracy (\%) is shown in the rightmost column. Highest values in each column are bolded.}
\label{tab:model_retrieval_comparison}
\resizebox{\textwidth}{!}{
\begin{tabular}{lccc|ccc|ccc|c}
\toprule
\textbf{Model} & \multicolumn{3}{c}{\textbf{Image$\rightarrow$Text}} & \multicolumn{3}{c}{\textbf{3D Scan$\rightarrow$Text}} & \multicolumn{3}{c}{\textbf{Text$\rightarrow$Image}} & \textbf{Linear Acc.} \\
\cmidrule(lr){2-4} \cmidrule(lr){5-7} \cmidrule(lr){8-10} \cmidrule(lr){11-11}
 & R@1 & R@5 & R@10 & R@1 & R@5 & R@10 & R@1 & R@5 & R@10 &  \\
\midrule
BiomedCLIP & 1.4 & 5.0 & 8.4 & 2.5 & 9.8 & 15.0 & 3.6 & 9.5 & 13.1 & 39.0 \\
BiomedCLIP (Fine-tuned) & 50.0 & 78.5 & 82.6 & 67.4 & 89.1 & 92.1 & 38.5 & 65.8 & 71.8 & 75.5 \\
ViT-B/16 (InfoNCE Loss) & 65.6 & \textbf{85.2} & \textbf{90.4} & 68.8 & 92.2 & 94.4 & 49.3 & 69.3 & 76.6 & 71.3 \\
ViT-S/16 & 46.7 & 79.1 & 84.4 & 69.0 & 92.2 & 95.2 & 64.6 & 77.8 & 80.9 & 73.6 \\
ViT-B/16 (\textbf{MR-CLIP})  & \textbf{66.0} & 77.3 & 78.3 & \textbf{78.7} & \textbf{94.2} & \textbf{95.3} & \textbf{90.9} & \textbf{93.6} & \textbf{94.4} & \textbf{82.6} \\
\bottomrule
\end{tabular}
}
\end{table*}

\begin{table*}[t]
\centering
\caption{Retrieval performance (\%) across different discretization granularities of TE and TR (TExTR). Top: results when training and evaluating on the same set. Bottom: generalization performance when transferring from a 20×20 training grid to other discretizations. We report R@1/5/10 for image-to-text, 3D scan-to-text, and text-to-image retrieval. Highest values in each column are bolded.}
\resizebox{\textwidth}{!}{
\begin{tabular}{l @{\hspace{0.4em}} ccc @{\hspace{0.4em}}|@{\hspace{0.4em}} ccc @{\hspace{0.4em}}|@{\hspace{0.4em}}ccc}
\toprule
\textbf{Set (\# total classes)} & \multicolumn{3}{c}{\textbf{Image→Text}} & \multicolumn{3}{c}{\textbf{3D Scan→Text}} & \multicolumn{3}{c}{\textbf{Text→Image}} \\
\cmidrule(lr){2-4} \cmidrule(lr){5-7} \cmidrule(lr){8-10}
 & \multicolumn{1}{c}{R@1} & \multicolumn{1}{c}{R@5} & \multicolumn{1}{c}{R@10} & \multicolumn{1}{c}{R@1} & \multicolumn{1}{c}{R@5} & \multicolumn{1}{c}{R@10} & \multicolumn{1}{c}{R@1} & \multicolumn{1}{c}{R@5} & \multicolumn{1}{c}{R@10} \\
\midrule
\multicolumn{10}{l}{\textit{Trained and evaluated on same sets}} \\
\midrule
40x20 (1770)    & 62.5 & 72.1 & 73.2 & 73.9 & 92.8 & 94.1 & 88.6 & 92.4 & 93.5 \\
20x20 (1415)    & 66.0 & 77.3 & 78.3 & 78.7 & 94.2 & 95.2 & 90.9 & 93.6 & 94.4 \\
20x10 (1017)    & 69.5 & 80.4 & 81.1 & 82.0 & 95.4 & 96.0 & 94.1 & 96.1 & 96.8 \\
10x10 (792)     & 77.2 & 84.3 & 85.2 & 86.4 & 96.1 & 96.6 & 92.0 & 96.5 & 97.6 \\
10x5 (599)      & 77.3 & 86.2 & 86.7 & 84.3 & 96.7 & 97.1 & 86.6 & 97.7 & 98.0 \\
5x5 (488)       & 69.7 & 86.9 & 88.0 & 89.1 & 97.0 & 97.4 & \textbf{93.7} & \textbf{97.8} & \textbf{98.1} \\
k-means(nc=100) (1220)  & 60.1 & 70.5 & 71.6 & 72.6 & 93.6 & 95.2 & 91.2 & 93.5 & 94.5 \\
k-means(nc=50) (826)  & 71.8 & 83.0 & 83.7 & 83.9 & 96.1 & 96.9 & 93.6 & 95.6 & 96.0 \\
k-means(nc=20) (522)  & \textbf{81.4} & \textbf{90.1} & \textbf{90.4} & \textbf{91.3} & \textbf{97.2} & \textbf{97.6} & 88.7 & 97.7 & 97.8 \\

\midrule
\multicolumn{10}{l}{\textit{Trained on 20x20 set, evaluated on other sets}} \\
\midrule
40x20 (1770)    & 54.1 & 74.1 & 76.7 & 68.0 & 93.2 & 94.7 & 74.1 & 84.4 & 88.4 \\
20x10 (1017)    & 71.1 & 80.1 & 80.8 & 81.7 & 95.4 & 96.1 & 93.2 & 95.2 & 95.7 \\
10x10 (792)     & 76.7 & 84.7 & 85.3 & 85.9 & 96.1 & 96.6 & 93.8 & 95.7 & 96.3 \\
10x5 (599)      & 78.9 & 86.2 & 86.6 & 87.3 & 96.6 & 97.1 & 94.7 & 96.5 & 97.0 \\
5x5 (488)       & \textbf{83.6} & \textbf{89.9} & \textbf{90.3} & \textbf{90.5} & \textbf{97.2} & \textbf{97.5} & \textbf{95.6} & \textbf{96.8} & \textbf{97.3} \\
\bottomrule
\end{tabular}
}
\label{tab:grid_transfer}
\end{table*}

\begin{table*}[b!]
\centering
\caption{Cross-modal retrieval performance for out-of-distribution dataset (\%). Showing R@1/5/10 for image-to-text, 3D scan-to-text, and text-to-image retrieval.}
\label{tab:model_retrieval_comparison_ood}
\resizebox{\textwidth}{!}{
\begin{tabular}{lccc|ccc|ccc}
\toprule
\textbf{Dataset} & \multicolumn{3}{c}{\textbf{Image$\rightarrow$Text}} & \multicolumn{3}{c}{\textbf{3D Scan$\rightarrow$Text}} & \multicolumn{3}{c}{\textbf{Text$\rightarrow$Image}} \\
\cmidrule(lr){2-4} \cmidrule(lr){5-7} \cmidrule(lr){8-10}
 & R@1 & R@5 & R@10 & R@1 & R@5 & R@10 & R@1 & R@5 & R@10 \\
\midrule
OASIS & 28.6 & 60.1 & 72.8 & 13.4 & 66.4 & 80.7 & 32.4 & 43.2 & 48.6 \\
OASIS (Only numerical tags) & 11.2 & 56.7 & 75.2 & \textbf{31.4} & \textbf{80.6} & \textbf{91.4} & 17.2 & 34.5 & 51.7 \\
OASIS (Labeled by numerical tags) & \textbf{43.3} & \textbf{69.9} & \textbf{80.9} & 19.0 & \textbf{80.6} & \textbf{91.4} & \textbf{37.8} & \textbf{54.1} & \textbf{59.5} \\
\bottomrule
\end{tabular}
}
\end{table*}
\textbf{Grouping Strategy of Numerical Tags:} We next study how the discretization of the TE-TR-TI space impacts retrieval performance. Table~\ref{tab:grid_transfer} presents results for varying grid sizes used to group acquisition parameters, both when training and testing on the same set (top) and when evaluating generalization from a fixed grid to other discretizations (bottom). We observe that coarser groupings (i.e., fewer total classes) generally lead to improved retrieval performance. For example, training on a 5$\times$5 TE-TR grid yields the best results across nearly all metrics, with R@10 reaching 88.0\% for image-to-text and 97.4\% for 3D scan-to-text retrieval. This suggests that reducing label granularity simplify the alignment task, potentially by reducing label noise and increasing intra-class coherence. We also compare our grid-based discretizations with k-means clustering, a data-driven approach tailored to the current dataset. While k-means with 20 clusters outperforms the grid method with closest number of classes (5x5) on image-to-text and 3D scan-to-text retrieval, it underperforms on text-to-image retrieval. Moreover, increasing the number of clusters results in slightly worse performance compared to their closest grid counterparts. This sensitivity highlights a key limitation of k-means: it requires manual tuning of the number of clusters (nc) and is tightly coupled to the training data distribution. Therefore, k-means-based grouping may struggle to generalize to different acquisition protocols or clinical settings. In contrast, the grid-based approach, though less optimal on some tasks, is likely to generalize better across diverse clinical settings.

Furthermore, to evaluate the adaptability of MR-CLIP to different use cases, we train it on a 20×20 TE-TR grid and test transfer performance on other discretizations without retraining. Despite training on finer labels, MR-CLIP showed strong performance on coarser labels, demonstrating its flexibility across varying contrast groupings and use cases.

\textbf{Generalizability:} We assess MR-CLIP’s ability to generalize beyond the training distribution by evaluating cross-modal retrieval on the OASIS dataset \cite{oasis}, which features differing acquisition characteristics. As shown in Table~\ref{tab:model_retrieval_comparison_ood}, using all available DICOM metadata yields reasonable performance (e.g., 60.1\% R@5 for image-to-text). In the second configuration, only numerical tags are used as input, as some categorical tags are either missing or not seen during training. This setup achieves comparable overall performance and improves 3D scan-to-text retrieval (e.g., 80.6\% R@5), highlighting the strength of model's reliance on numerical metadata alone. The third setup retains categorical tags as input but uses only numerical tags for label construction. Even when incomplete or previously unseen, categorical metadata provides complementary information, leading to improved performance across tasks (e.g., 69.9\% R@5 for image-to-text and 54.1\% R@5 for text-to-image).

\textbf{Error Rates Across DICOM Tags:} We evaluate the performance of our model on individual DICOM metadata tags by measuring the percentage of mismatches between the predicted and ground truth values using linear-probe classification. Among these, discretized scan parameters such as TE and TR show relatively high classification error rates—16.49\% and 11.65\%, respectively, even though the average prediction is close to the true value. Specifically, the average bin-level errors for TE and TR are only 1.44 and 2.29 bins, which correspond to approximate deviations of 14.4 ms in TE and 1145 ms in TR (based on a 20×20 TE–TR grid). This suggests that while predictions often land near the correct value, they are not always in the exact same bin, likely due to the semantic ambiguity introduced by discretization. In contrast, clearly defined tags such as \textit{Field Strength} show near-zero error, indicating strong model reliability.


\begin{figure*}[t!]
\centering
\includegraphics[width=0.7\textwidth]{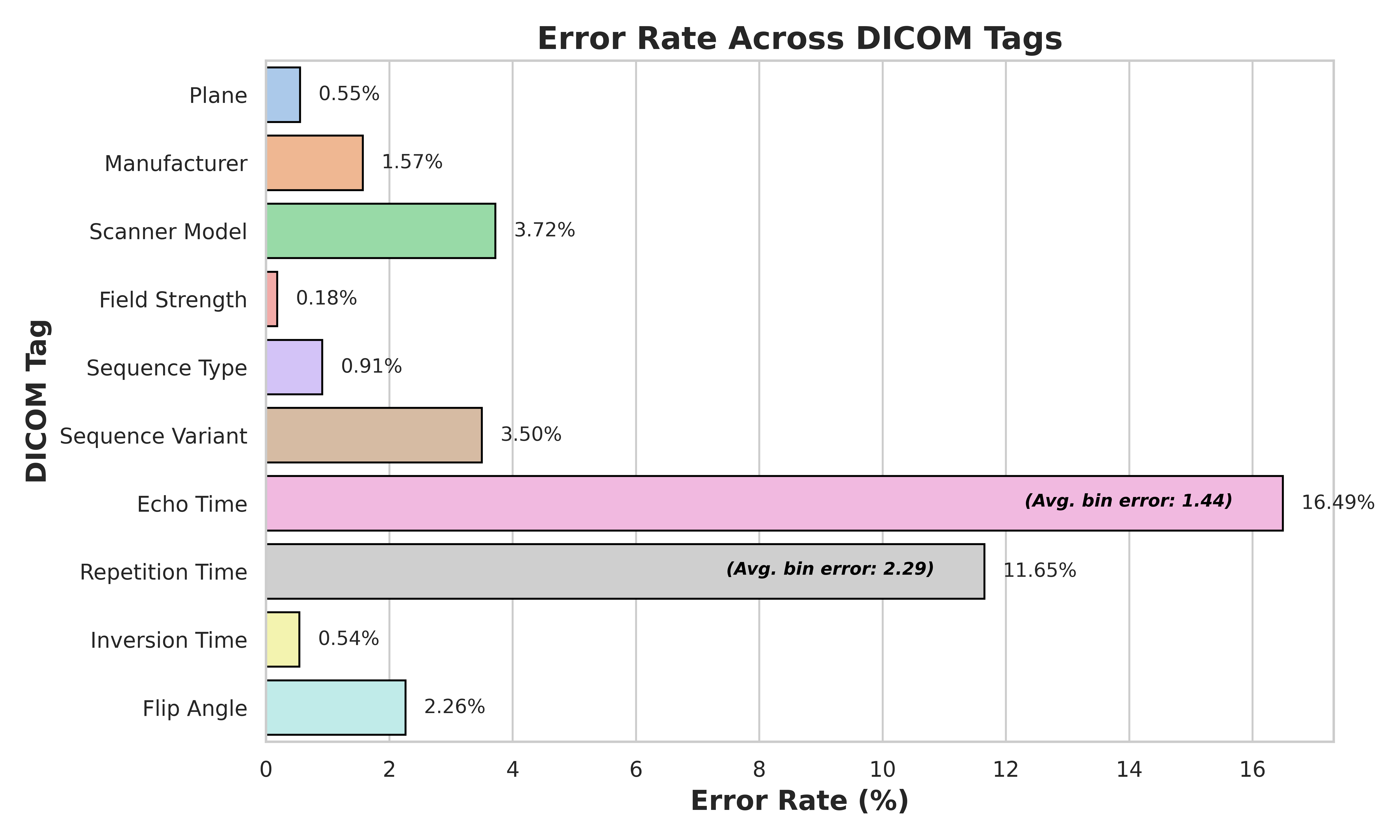}
\caption{Error rates across DICOM tags based on linear probe classification results. 
}
\label{fig3}
\end{figure*}
\section{Discussion and Conclusion}
We introduced MR-CLIP, a multimodal contrastive learning framework that aligns MR images with their DICOM metadata to learn robust, contrast-aware representations without manual annotations. By converting acquisition parameters into structured text prompts and using a supervised contrastive loss, MR-CLIP effectively decouples contrast information from anatomical content, enabling generalization across scanners, protocols, and anatomical regions. This makes it especially suited for the heterogeneity of real-world clinical data. While MR-CLIP shows strong potential, it relies on the assumption that available DICOM metadata reflects relevant contrast information accurately and comprehensively. We note that certain imaging variants may not be fully captured by the selected fields, and metadata errors due to scanner inconsistencies or manual entry mistakes (e.g. series description) can introduce noise into the learning process. Addressing these limitations will require validation against expert annotations, and exploring automated approaches for identifying erroneous metadata.

Future work will extend MR-CLIP to native 3D data and additional modalities (e.g CT, PET), aiming towards an unified, modality-agnostic representation space. We intend to explore conditioning on patient-specific metadata for personalized retrieval and integrate domain adaptation techniques, including cross-site and cross-modality data harmonization. These directions will advance MR-CLIP towards a scalable, foundation model for medical imaging, with broad applicability across retrieval, harmonization, and clinical decision support.

\begin{credits}
\subsubsection{\ackname} This work was supported by the UK Engineering and Physical Sciences Research Council (EPSRC) [Grant reference number EP/Y035216/1] Centre for Doctoral Training in Data-Driven Health (DRIVE-Health) at King's College London, with additional support from deepc GMBH.

\subsubsection{\discintname}
M.Y. Avci's PhD is partially funded by deepc GMBH.
\end{credits}
%
%
%
%
%
\bibliographystyle{splncs04}
\bibliography{main}

\begin{thebibliography}{10}
\providecommand{\url}[1]{\texttt{#1}}
\providecommand{\urlprefix}{URL }
\providecommand{\doi}[1]{https://doi.org/#1}

\bibitem{contextmri}
Chung, H., Lee, D., Wu, Z., Kim, B.H., Bouman, K.L., Ye, J.C.: Context{M}{R}{I}: Enhancing compressed sensing {M}{R}{I} through metadata conditioning. arXiv preprint arXiv:2501.04284  (2025)

\bibitem{du2020radnet}
Du, R., Vardhanabhuti, V.: 3{D}-{R}{A}{D}{N}et: Extracting labels from {D}{I}{C}{O}{M} metadata for training general medical domain deep 3{D} convolution neural networks. In: Proceedings of the Third Conference on Medical Imaging with Deep Learning. vol.~121, pp. 174--192. PMLR (2020)

\bibitem{gauriau2020dicom}
Gauriau, R., Bridge, C., Chen, L., Glocker, B., Hajnal, J., Rueckert, D., Bai, W.: Using {D}{I}{C}{O}{M} metadata for radiological image series categorization: A feasibility study on large clinical brain {M}{R}{I} datasets. Journal of Digital Imaging  \textbf{33}(3),  747--762 (2020)

\bibitem{oct-clip}
Holland, R., Leingang, O., Bogunović, H., Riedl, S., Fritsche, L., Prevost, T., Scholl, H.P., Schmidt-Erfurth, U., Sivaprasad, S., Lotery, A.J., Rueckert, D., Menten, M.J.: Metadata-enhanced contrastive learning from retinal optical coherence tomography images. Medical Image Analysis  \textbf{97},  103296 (2024)

\bibitem{synthstrip}
Hoopes, A., Mora, J.S., Dalca, A.V., Fischl, B., Hoffmann, M.: Synthstrip: skull-stripping for any brain image. NeuroImage  \textbf{260},  119474 (2022)

\bibitem{open_clip}
Ilharco, G., Wortsman, M., Wightman, R., Gordon, C., Carlini, N., Taori, R., Dave, A., Shankar, V., Namkoong, H., Miller, J., Hajishirzi, H., Farhadi, A., Schmidt, L.: Openclip (Jul 2021)

\bibitem{supcon}
Khosla, P., Teterwak, P., Wang, C., Sarna, A., Tian, Y., Isola, P., Maschinot, A., Liu, C., Krishnan, D.: Supervised contrastive learning. In: Proceedings of the 34th International Conference on Neural Information Processing Systems. NIPS '20, Curran Associates Inc., Red Hook, NY, USA (2020)

\bibitem{oasis}
LaMontagne, P.J., Benzinger, T.L., Morris, J.C., Keefe, S., Hornbeck, R., Xiong, C., Grant, E., Hassenstab, J., Moulder, K., Vlassenko, A.G., Raichle, M.E., Cruchaga, C., Marcus, D.: {O}{A}{S}{I}{S}-3: Longitudinal neuroimaging, clinical, and cognitive dataset for normal aging and {A}lzheimer disease. medRxiv  (2019)

\bibitem{liang2021mri}
Liang, S., Beaton, D., Arnott, S.R., Gee, T., Zamyadi, M., Bartha, R., Symons, S., MacQueen, G.M., Hassel, S., Lerch, J.P., Anagnostou, E., Lam, R.W., Frey, B.N., Milev, R., Müller, D.J., Kennedy, S.H., Scott, C.J.M., {ONDRI Investigators}, Strother, S.C.: Magnetic {R}esonance {I}maging sequence identification using a metadata learning approach. Frontiers in Neuroinformatics  \textbf{15},  622951 (2021)

\bibitem{mcdaniel2022contrast}
McDaniel, J.W., Moore, A.Z., Mareci, T.H., Price, S.L., Conklin, D.J., Wagle, N., Pinho, M.C., Landman, B.A.: Improving the automatic classification of brain {M}{R}{I} acquisition contrast with machine learning. Academic Radiology  (2022)

\bibitem{DICOMStandard}
{National Electrical Manufacturers Association}: Digital imaging and communications in medicine ({D}{I}{C}{O}{M}) standard. \url{https://www.dicomstandard.org/} (May 2025), nEMA PS3 / ISO 12052

\bibitem{infonce}
van~den Oord, A., Li, Y., Vinyals, O.: Representation learning with contrastive predictive coding. arXiv preprint arXiv:1807.03748  (2019)

\bibitem{ouyang2021representation}
Ouyang, J., Adeli, E., Pohl, K.M., Zhao, Q., Zaharchuk, G.: Representation disentanglement for multi-modal brain {M}{R}{I} analysis. In: Information Processing in Medical Imaging (IPMI). Lecture Notes in Computer Science, vol. 12729, pp. 321--333. Springer (2021)

\bibitem{radford2021learning}
Radford, A., Kim, J.W., Hallacy, C., Ramesh, A., Goh, G., Agarwal, S., Sastry, G., Askell, A., Mishkin, P., Clark, J., Krueger, G., Sutskever, I.: Learning transferable visual models from natural language supervision. In: Proceedings of the International Conference on Machine Learning (2021)

\bibitem{sinha2024mrqa}
Sinha, H., Raamana, P.R.: Solving the pervasive problem of protocol non-compliance in {M}{R}{I} using an open-source tool mr{Q}{A}. Neuroinformatics  \textbf{22},  297--315 (2024)

\bibitem{wang2022medclip}
Wang, H., Liu, K., Ng, N., et~al.: Med{C}{L}{I}{P}: Contrastive learning from unpaired medical images and text. In: International Conference on Medical Image Computing and Computer-Assisted Intervention (2022)

\bibitem{tumsyn}
Wang, Y., Xiong, H., Sun, K., Bai, S., Dai, L., Ding, Z., Liu, J., Wang, Q., Liu, Q., Shen, D.: Towards general text-guided image synthesis for customized multimodal brain {M}{R}{I} generation. arXiv preprint arXiv:2409.16818  (2024)

\bibitem{wu2023pmcclip}
Wu, J., Zhang, Y., Xie, Y., Ma, T., et~al.: {PMC-CLIP}: Contrastive vision-language pretraining on biomedical literature. In: Proceedings of the IEEE/CVF Conference on Computer Vision and Pattern Recognition (2023)

\bibitem{biomedclip}
Zhang, S., Xu, Y., Usuyama, N., Xu, H., Bagga, J., Tinn, R., Preston, S., Rao, R., Wei, M., Valluri, N., Wong, C., Tupini, A., Wang, Y., Mazzola, M., Shukla, S., Liden, L., Gao, J., Crabtree, A., Piening, B., Bifulco, C., Lungren, M.P., Naumann, T., Wang, S., Poon, H.: A multimodal biomedical foundation model trained from fifteen million image–text pairs. NEJM AI  \textbf{2}(1) (2024)

\bibitem{haca3}
Zuo, L., Liu, Y., Xue, Y., Dewey, B.E., Remedios, S.W., Hays, S.P., Bilgel, M., Mowry, E.M., Newsome, S.D., Calabresi, P.A., Resnick, S.M., Prince, J.L., Carass, A.: {HACA3}: A unified approach for multi-site {MR} image harmonization. Computerized Medical Imaging and Graphics  \textbf{109},  102285 (2023)

\end{thebibliography}
\end{document}